\documentclass[twoside,11pt]{article}

\usepackage{jmlr2e}

\usepackage[utf8]{inputenc} 
\usepackage[T1]{fontenc}    

\newcommand{\footremember}[2]{%
   \thanks{#2}
    \newcounter{#1}
    \setcounter{#1}{\value{footnote}}%
}
\newcommand{\footrecall}[1]{%
    \footnotemark[\value{#1}]%
}

\hypersetup{
    colorlinks=true,
    linkcolor=blue,
    filecolor=magenta,      
    urlcolor=cyan,
}
\usepackage{url}            
\usepackage{booktabs}       
\usepackage{amsfonts}       
\usepackage{nicefrac}       
\usepackage{microtype}      
\usepackage{tabu}
\usepackage {tikz}
\usetikzlibrary {positioning}
\usepackage{amssymb}
\usepackage{amsmath}
\usepackage{amsfonts}
\usepackage{graphicx}
\usepackage{filecontents}
\usepackage{subcaption}

\usepackage{times}
\usepackage{epsfig}
\usepackage{graphicx}
\usepackage{amssymb}
\usepackage{mathtools}
\usepackage{listings}
\usepackage{float}
\usepackage{subcaption}
\usepackage{mwe}
\usepackage{ltablex}
\usepackage{siunitx}
\usepackage{caption}%
\usepackage[flushleft]{threeparttablex}
\usepackage{adjustbox}

\usepackage{stfloats}
\usepackage{lipsum}

\ShortHeadings{Deep EHR: Chronic Disease Prediction Using Medical Notes}{Liu, Zhang and Razavian}
\firstpageno{1}

\begin{document}

\title{Deep EHR: Chronic Disease Prediction Using Medical Notes}

\author{\name Jingshu Liu\footremember{note1}{Equal contribution} \email jingshu.liu@nyu.edu \\
       \addr New York University
       \AND
       \name Zachariah Zhang\footrecall{note1} \email zz1409@nyu.edu \\
       \addr New York University
       \AND
       \name Narges Razavian \email narges.razavian@nyumc.org \\
       \addr New York University
       } 

\maketitle

\begin{abstract}
  Early detection of preventable diseases is important for better disease management, improved interventions, and more efficient health-care resource allocation. Various machine learning approaches have been developed to utilize information in Electronic Health Record (EHR) for this task. Majority of previous attempts, however, focus on structured fields and lose the vast amount of information in the unstructured notes. 
  In this work we propose a general multi-task framework for disease onset prediction that combines both free-text medical notes and structured information. We compare performance of different deep learning architectures including CNN, LSTM and hierarchical models. In contrast to traditional text-based prediction models, our approach does not require disease specific feature engineering, and can handle negations and numerical values that exist in the text. Our results on a cohort of about 1 million patients show that models using text outperform models using just structured data, and that models capable of using numerical values and negations in the text, in addition to the raw text, further improve performance. Additionally, we compare different visualization methods for medical professionals to interpret model predictions. 
  
\end{abstract}

\section{Introduction}

The past decade has witnessed a sky-rocketing increase of information in EHR systems. Structured patient information such as demographics, disease history, lab results, procedures and medications, and unstructured information such as progress notes and discharge notes are collected during each clinical encounter. This creates an opportunity to mine the information to increase the quality of care. Yet physicians have limited time to process all the available data for each patient, let alone to detect patterns across similar patients. Machine learning approaches, on the other hand, are suitable for extracting information from vast amount of data and generalizing to new cases.

Recent studies have shown promising results using EHR and deep learning models to predict clinical events. However, previous studies mainly focused on modelling with structured data such as lab results, clinical measurements [\cite{lipton2015learning},  \cite{razavian2016multi}] and historical diagnoses codes [\cite{choi2016doctor}]. \cite{sureshclinical} predicted clinical intervention combining structured data and notes. Yet the author resolved to transforming each clinical narrative note to a 50-dimensional vector of topic proportions with LDA. These methods lack capacity of extracting rich information from unstructured medical notes data. For example, the fact that a patient is brought in by their family members isn't coded but can be an indicator of a bad health as well as an existing social support for the patient. A more recent study, \cite{baumel2017multi}, attempted to identify ICD code assignment based on MIMIC discharge notes and showed that deep learning based methods outperforms shallower ones.   

In this study we present a general framework for predicting onset of diseases that emphasize the following: (1) Flexibility to \textbf{utilize both unstructured text and structured numerical values}. Vector representation of words or sequence of text, such as pre-trained embeddings, can be easily combined with other numerical data. (2) \textbf{Multi-task framework} that can be generalized to different diseases. We test model performance on the prediction of three disease areas, namely congestive heart failure, kidney failure and stroke. Since we do not require disease specific feature engineering or model structure, the same architecture can be readily leveraged to other disease areas. (3) Evaluation on a large cohort of real world patient data, with \textbf{various note types and note lengths}. 

We experiment with variants of deep learning model architectures, including recurrent neural network with Long Short-term Memory units (LSTM) and Convolutional Neural Networks (CNN). In addition we propose a novel technique for handling negations in this prediction task.

In terms of performance, we show that models using medical notes outperform those with only lab and demographic data. Furthermore, deep learning methods achieve better performance than logistic regression baseline with TF-IDF features. We find that several models, particularly a BiLSTM with negation tags and lab and demographic features, can achieve high AUC for all three diseases.

In order to help medical professionals to interpret model output, we further compare the effectiveness of several visualization methods to identify the words and phrases with relatively high impact on the model prediction. We find log-odds based approach provides more intuitive visualization while gradient based approach tends to be too noisy.

Modeling codes and the pre-trained embeddings are available at
\url{https://github.com/NYUMedML/DeepEHR}.

\section{Prediction Task}

\begin{figure}[h]
    \centering
    \includegraphics[width=.75\textwidth]{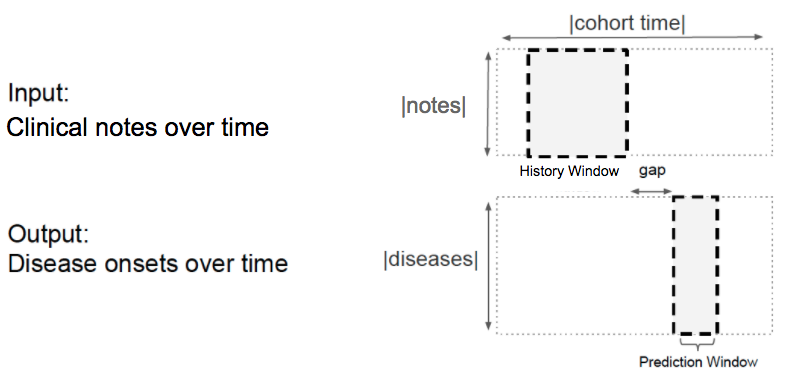}
    \caption{Overview of prediction framework}\label{fig:predTask}
\end{figure}

Figure \ref{fig:predTask} illustrates the prediction task. The goal is to forecast the onset of diseases prior to diagnosis time. We use information accumulated in the history window to predict diagnoses in the prediction window. We also require a gap period between the end of the history window and the beginning of the prediction window. The purpose is to prevent the model from cheating with information generated right before the diagnosis time. Specifically, we use a 12-month historical window, a 3-month gap and a 6-month prediction window in this study. 

\section{Cohort}
We use medical notes, demographics and diagnoses in ICD-10 codes from the NYU Langone Hospital EHR system. The data contains clinical encounters of more than 1 million patients between 2014 and 2017, and more than 15 million entries of medical notes. Figure \ref{fig:notetype1} and Figure \ref{fig:notetype2} in the Appendix show the distribution of note types and note lengths by type. Note lengths typically range from hundreds to thousands of words, though some notes are extremely long with more than 10K words. Section \ref{sec:Text} provides details on text preprocessing. 

Additionally, more than $60\%$ of encounters contain numerical information including vital signs and lab test values in the notes. We extract the most common ones from the text and use them as additional input to the model. More details on numerical value extraction are provided in Section \ref{sec:numValue}.

\subsection{Target Disease Definition}
We define the mapping between the ICD-10 codes and the three target diseases by a union of Centers for Medicare and Medicaid Services chronic conditions (CMS-CCW, \cite{CCW}) and the Healthcare Cost and Utilization Project Clinical Classifications Software (HCUP-CCS, \cite{HCUP}), with help from medical experts from NYU Medical School. Detailed mapping can be found in the Appendix \ref{sec:icd10}.


\subsection{Modeling Dataset}

We limit the analysis to patients with at least two encounters (of any kind) in the historical window and at least two encounters (of any kind) in the prediction window. We further require the gap between the first and last encounters in the prediction window to be longer than or equal to two months, to avoid the case of patient switching hospitals and getting diagnosed elsewhere. Moreover, patients with the target diagnosis in any encounters prior to the prediction window are excluded, to prevent modeling patients who already have the target disease. We utilize a masking approach to exclude patients that are valid examples for one disease but may be invalidated for another (implementation details described in Section \ref{sec:multitask}). For example, a patient might already have heart failure but we would like to predict the onset of kidney failure or stroke. The modeling cohort thus reduces to about 300k patients.

For patients in the modeling cohort, we observe their encounters up to 3 years. As the total length of the history window, the gap period and the prediction window is less than 2 years, we use a 3-month sliding window to utilize all of the data. As a result, the same patient may contribute multiple data points. We then split the modeling data into training, validation and test set by a ratio of 7:1:2 by patient ID. The same patient never appears in two separate sets. Table \ref{table:data} shows the number of negative cases versus positive cases for each disease. 

\begin{table}[h]
    \centering
    \caption{Number of Records by Target Diseases (Negative Cases : Positive Cases)}
\begin{tabular}{ l  c  c  c }
\hline
Target & Training Set & Validation Set & Test Set \\ \hline
 Congestive Heart Failure & 644K : 4080 & 93K : 574 & 184K : 1167 \\ 
 Kidney Failure & 616K : 10051 & 88K : 1428 & 176K : 2809 \\ 
 Stroke & 653K : 3195 & 94K : 406 & 187K : 916 \\ \hline
\end{tabular}

\label{table:data}
\end{table}

\subsection{Data Modalities} 

In our models we consider the following three types of input data: raw text of medical notes, lab and vital sign data recovered from medical notes, as well as structured demographic data.

\subsubsection{Text} 
\label{sec:Text}
 We remove stop words from the text and select a vocabulary of the 20k most frequent words. Words that appear in more than $80\%$ of documents are also ignored. In addition, we replace numbers (after numerical value extraction as described in the following section),
  deidentified names, addresses, and locations with generic tokens. On average each record contains around 1,300 words after preprocessing, and the 90th percentile is around 3,000.
 
 We represent words as dense word embeddings, in combination with downstream convolution and recurrent architectures as discussed in detail in Section \ref{sec:methods}. We experiment with two types of word embedding, one trained on the PubMed dataset by \cite{moen2013distributional} , and another trained on our data and optimized for our task trained with StarSpace [\cite{wu2017starspace}] method. More details are presented in section \ref{sec:ssp}.

\subsubsection{Numerical Lab and Vital Sign Values} 
\label{sec:numValue}

We adopt and modify Valx by \cite{hao2016valx} to extract numerical lab and vital sign values from the note text. Valx is a simple regular expression and heuristic rule-based tool to identify test names according to target dictionaries, and associate numerical values to their test names. In terms of target dictionary, we include common vital signs (i.e., weight, height, BMI, blood pressure, temperature, pulse and respiration rate) as well as lab test names based on the CDISC STDM terminology [\cite{cdisc}]. For simplicity we will refer to both vital signs and lab test values as lab values in the following sections.

Overall more than 700 types of lab test are extracted from the notes. The most frequent item, weight, is found in about $65\%$ of encounters, while the prevalence drops quickly to around $1\%$ at the $50$th most frequent item. Table \ref{table:labs} in the appendix shows the top 30 most frequent lab tests and their prevalence by percentage of encounters in the training set.  

In each encounter, the same item may have multiple values if the patient was tested multiple times in a day. We calculate the median, min and max within each encounter. We also normalized the values by subtracting mean and dividing by the standard deviation of the training set distribution. Missing values are imputed with $0$ if no previous test results exist for the same patient. Otherwise we carry over the previous results to later encounters until a new result is observed. Values of the top 50 most frequent lab tests are included in the model as vectors with length 150. 

\subsubsection{Demographics} 
We use structured data on patients ethnicity, race, gender, and age from the EHR system as 
demographic features have been shown to be important for disease prediction. Ethnicity, race, and gender are represented as one-hot encoded categorical features. Age is represented as a continuous variable in years. This yields a total of 61 additional features. Feature distributions are provided in the appendix figures \ref{fig:distethnicity}, \ref{fig:distrace} and \ref{fig:distgender}. Demographic features are available for about $90\%$ of patients.

\section{Methods}

\subsection{Baselines}
We use three different baselines to evaluate the contribution of text data, the deep learning architecture and their combination. Firstly, we train an L1-regularized logistic regression model with all available demographic features and lab values averaged within the history window. Secondly we train an LSTM model with all available demographic features and lab values at each encounter, to provide a more fair comparison with other deep learning models with additional text data. The last baseline is an L1-regularized logistic regression with TF-IDF N-gram features in the text. This baseline shows the value of deep learning in modeling complicated text. We use the 20k most frequent 1-, 2-, and 3-grams. We utilize the sklearn's implementation of logistic regression for this task \cite{scikit-learn}.

\subsection{Learning Continuous Embedding of Vocabulary}
\label{sec:ssp}

\begin{figure}[h]
    \centering
    \includegraphics[width=.95\textwidth]{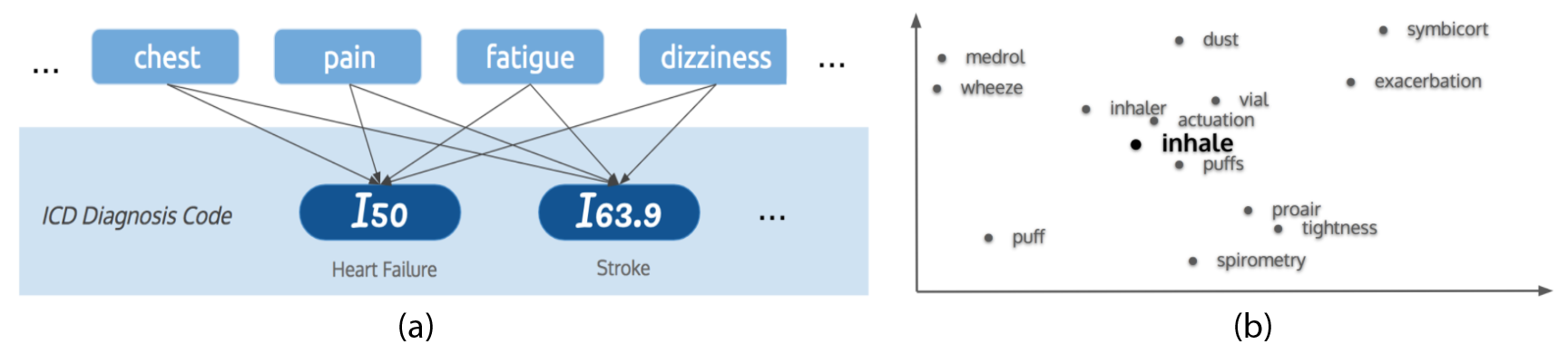}
    \caption{Illustration of StarSpace: (a) embeddings are trained with labeled bag-of-words approach. (b) StarSpace word embedding example under t-SNE projection. }\label{fig:starspace}
\end{figure}

In our first analysis we use embeddings previously trained on the PubMed dataset \cite{moen2013distributional}. We then train new embeddings directly on the NYU Langone Center medical notes, as the style and abbreviations present in clinical notes are distinct from medical publications available at PubMed. We adopt StarSpace \cite{wu2017starspace} as a general-purpose neural model for efficient learning of entity embeddings. In particular, we label the notes from each encounter with the ICD-10 diagnosis codes of the same encounter, as shown in Figure \ref{fig:starspace} (a). Under StarSpace's bag-of-word approach, the encounter is represented by aggregating the embedding of individual words (we used the default aggregation method where the encounter is the sum of embeddings of all words divided by the squared root of number of words). Both the word embeddings and the diagnosis code embeddings are trained so that the cosine similarity between the encounter and its diagnoses is ranked higher than that between the encounter and a set of different diagnoses. Thus words related to the same symptom are placed close to each other in the embedding space. For example, figure \ref{fig:starspace} (b) shows neighbours of the word "inhale" by t-SNE projection of the embeddings to the 2-dimensional space. We find that the bag-of-word style embedding creates representations better for disease prediction than using a standard skip-gram objective.

Utilizing all available notes except for those on patients in the validation or the test set, we obtain a much larger embedding training set than that of the prediction task. We find that the StarSpace embeddings trained on clinical notes outperform the pre-trained PubMed embeddings in the downstream prediction task, as shown in Table \ref{tab:results}. We test a set of 300-dimension embeddings and 50-dimension ones early in the process and found the former outperformed the latter in the prediction tasks. We use the 300-dimension ones as input for all the deep learning models discussed in later sections. 

\subsection{Deep Learning Models} \label{sec:methods}

We explore various successful architectures including CNN, LSTM and models with different hierarchical structures. We train each model with pre-trained StarSpace embeddings as described in Section \ref{sec:ssp}. We do not update the embeddings during training of prediction models for training speed considerations. 




\subsubsection{Convolutional Neural Network (CNN)}
\label{sec:CNN}

\begin{figure}[h]
\begin{center}
\includegraphics[width=0.9\linewidth]{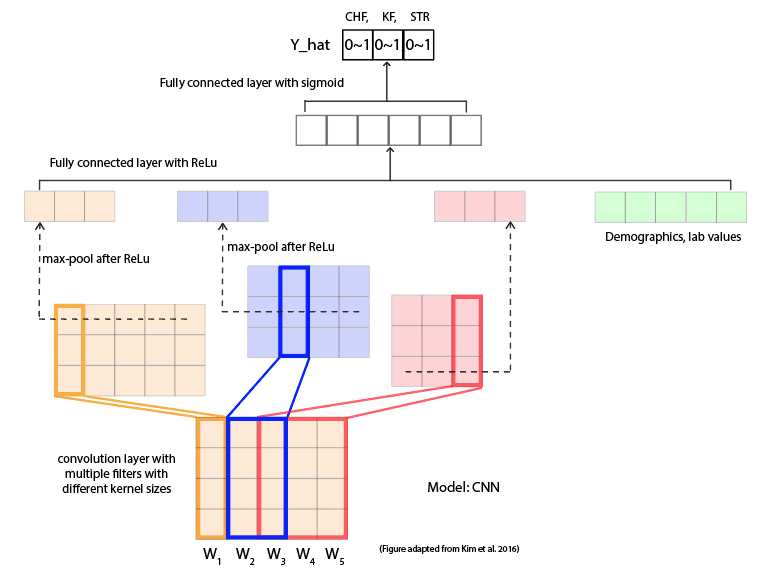}
\end{center}
   \caption{Illustration of Convolutional Neural Network model architecture: 1-D convolutions are performed on the entire input sequence with various kernel sizes, max-pooled over time and then concatenated. Lab values and demographics are concatenated to the same vector, then fed through fully connected layers.}
\label{fig:CNN}
\end{figure}

The Convolutional Neural Network model is based on the architecture in \cite{kim2014convolutional}. This model concatenates representations of the text at different levels of abstraction, essentially chooses the most salient n-grams from the notes. We perform 1D convolutions on the input sequence with kernel sizes 1, 2, and 3. The outputs of each of these convolutions are max-pooled over time and then concatenated. This representation is fed through one or more fully connected layers and a sigmoid output layer to make a prediction. We find that adding a dense hidden layer after max-pooling adds predictive value as it is able to learn interactions between the text features. We concatenate all encounter notes in the history period end to end as input to this network. We integrate lab and demographic features by concatenating the mean of each feature within the history window to the hidden states after the max pooling layer. Figure \ref{fig:CNN} illustrates the model architecture.

\subsubsection{Recurrent Neural Networks: Long Short Term Memory(LSTM) and Bidirectional LSTM (BiLSTM) }

\begin{figure}[h]
\begin{center}
\includegraphics[width=0.9\linewidth]{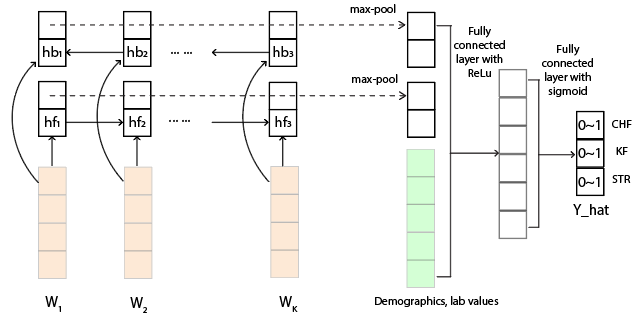}
\end{center}
   \caption{Illustration of the Bi-LSTM model architecture: hidden states are generated by processing one word at a time, then max-pooled across time. Lab values and demographics are concatenated to the output from Bi-LSTM and then fed into fully connected layers.}
\label{fig:BiLSTM}
\end{figure}

LSTM [\cite{hochreiter1997long}] models are variants of Recurrent Neural Networks with memory gates that take a single input word at each time step and update the models' internal representation accordingly. LSTM models have been used extensively for sequence learning problems, particularly in natural language processing domain. BiLSTM model [\cite{graves2013speech}] is an extension of the standard LSTM models, that combine two LSTMs with one running forward in time and the other running backward. Thus the context window around each word consists of both information prior to and after the current word. 

We experiment with LSTM and BiLSTM models directly on the word sequence of all the notes associated with each patient. However, the fact that our data contains generally very long notes creates a challenge for preserving the gradient across thousands of words. We initially experimented with target replication [\cite{target_rep}] to alleviate this problem. This technique predicts the presence of disease at every time step in order to model longer sequences. However, we found that it did not improve our model performance for any of the diseases. 

As an alternative, we explore two different approaches, max pooling across the BiLSTM hidden states as shown in Figure \ref{fig:BiLSTM}, as well as an encounter-level hierarchical model as described in section \ref{sec:enc_hier}. In the BiLSTM with max pooling architecture, similar to the Convolutional architecture, lab and demographic features are concatenated to the feature output from the BiLSTM. 

\subsubsection{Encounter Level Hierarchy}
\label{sec:enc_hier}
\begin{figure}[h]
\begin{center}
\includegraphics[width=0.95\linewidth]{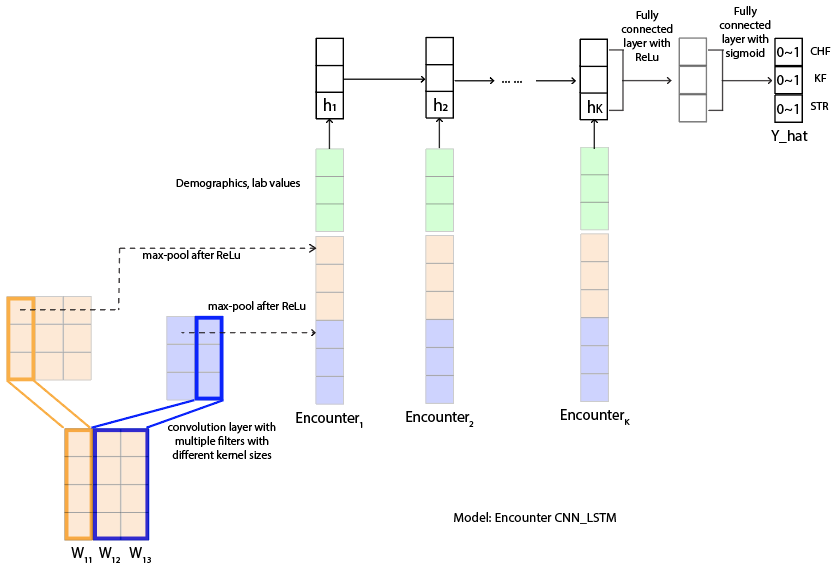}
\end{center}
   \caption{Illustration of the encounter-level hierarchy model: each encounter is encoded into a hidden vector via a CNN. Lab values and demographics are concatenated to the CNN output, then fed into an LSTM to generate prediction.}
\label{fig:Enc}
\end{figure}

Another approach is to add hierarchy at the encounter level. In this model we represent all encounters as a sequence over the history window. The motivation includes both reducing word sequence length and to capture momentum in a patient's health. For example, if notes of a patient show more and more heart disease related terms, we would expect him/her to be at a higher risk for heart disease than those with less or no such terms in later notes. 
This model first encodes each encounter into hidden states $h_0, h_1, ... h_K$ using convolutions, similar to the architecture discussed in Section \ref{sec:CNN}. Each of these states is then fed into an LSTM. The final hidden state is used for predictions. Here we incorporate numerical values by concatenating the feature vector to the CNN encoder output at each encounter. Demographic values are repeated across encounters. Figure \ref{fig:Enc} illustrates the model architecture.

\subsection{Multitask Learning}
\label{sec:multitask}
Multitask learning has shown to be an effective method of regularization in many deep learning applications. We compare the multitask approach with training each model separately with the CNN model. As shown in Table \ref{tab:results}, CNN outperforms CNN Single Task for all three diseases.

The multitask version of the models simply includes multiple sigmoid outputs at the final layer, one for each disease. Since a patient can be a valid data point for one disease but not another, we implement a masked binary cross entropy loss. Patients with diagnosis of one disease prior to the prediction window are masked from loss calculation for the corresponding disease, to prevent learning from patients not in a disease's cohort.

\subsection{Negation Tagging}

The notes are full of examples of patient having or not having certain conditions, making it very important to accurately interpret negations. For example, a patient having 'atrial fibrillation' is more likely to have heart failure and not having it is the opposite. CNNs tend to fail to capture this when the size of the kernel is smaller than the length of the negation sequence. With longer sequence, LSTMs can also suffer from similar issues.

To address this problem we tag all negated phrases through a simple prepossessing step. We then take the negative of the original word embedding as the embedding of each negated token. To tag negations, we use the Negex system[Chapman \textit{et al.} \cite{chapman2001simple}], which is a regular expression based system optimized for tagging negation in medical text. We see consistent improvement of prediction score across all our architectures. Details of results are discussed in table \ref{tab:results}. Below is an example of how adding negation tags corrects a false positive for stroke prediction. Nevertheless, it should be noted that the accuracy of negation tagging is a bottleneck in this project and more value could potentially be added with a more sophisticated negation tagging model. \newline

\fbox{\begin{minipage}{39em}
Original note:

... no known allergies review of symptoms : general : no fevers , chills , or weight loss... no cough , shortness of breath , or wheezing cardiovascular : no chest pain or dyspnea on exertion gastrointestinal : no abdominal pain , change in bowel habits , or black or bloody stools... neurological : no transient ischemic attack or stroke symptoms...\newline

Negation Tagged:

... no known \textbf{allergies\_neg} review of symptoms : general : no \textbf{fevers\_neg} , \textbf{chills\_neg} , or \textbf{weight\_neg loss\_neg}... no \textbf{cough\_neg }, \textbf{shortness\_neg} of \textbf{breath\_neg} , or \textbf{wheezing\_neg} cardiovascular : no \textbf{chest\_neg pain\_neg} or \textbf{dyspnea\_neg} on \textbf{exertion\_neg} gastrointestinal : no \textbf{abdominal\_neg pain\_neg , change\_neg in\_neg bowel\_neg habits\_neg} , or black or bloody stools... neurological : no transient \textbf{ischemic\_neg attack\_neg} or \textbf{stroke\_neg symptoms\_neg}...\newline

Actual Outcome: \textbf{No Stroke}

Prediction without Tags:  \textbf{0.8583}

Prediction with Tags: \textbf{0.3285}

\end{minipage}}

\subsection{Additional Explorations}

In addition to the architectures discussed above, we test the following variations but find them to be unsuccessful: We attempt to enforce non-negativity in the classification layer weights of the network. The motivation for this is that we generally observe many weak negative predictors for each disease, and we suspect them acting as bias adjusters. We also attempt to include sentence level hierarchy instead of encounter level hierarchy, but find its performance inferior to existing architectures. We suspect that this architecture introduces many new parameters and the model is unable to learn any more meaningful features.

\section{Results} 
Details on the exact architecture hyper-parameters for each model are given in Appendix \ref{sec:architecture}. All models are optimized with Adam optimization with a learning rate of .001, on GPUs provided by NYU Langone Medical Center High Performance Computing. All non-hierarchical models are trained with a mini-batch size of 128 with sequences padded to 3000 tokens. The encounter hierarchical models are padded to 30 encounters and 800 tokens per encounter, and trained with a mini-batch size of 16. As classes are highly imbalanced in the data, we split the positive and negative cases in the training set and use balanced number of cases in each mini-batch.

 \subsection{Prediction Accuracy}

\begin{table}[h]
    \centering
    \caption{Model Performance (AUC) by Target Disease}
    \label{tab:results}
    \begin{tabular}{| l | m{2.5cm} | m{2.5cm} | m{2.5cm} |  }
        \hline
     & \textbf{Heart Failure} & \textbf{Kidney Failure} & \textbf{Stroke} \\ \hline
    Logistic Reg Lab/Demo & 0.781 & 0.724 & 0.70\\ \hline
    LSTM Lab/Demo & 0.813 & 0.743 & 0.699 \\ \hline
    Logistic Reg Notes & 0.810 & 0.752  &  0.708\\ \hline
    \hline
    CNN PubMed Embeddings & 0.844 &  0.799 & 0.711\\ \hline 
    CNN Single Task & 0.847 &  0.796  & 0.706\\ \hline
    CNN & 0.854 &  0.802  & 0.714\\ \hline
    CNN + Neg Tag 	& 0.867  & 0.811  & 0.727 \\ \hline
    CNN + Neg Tag + Dense 	& 0.880  & 0.812  & 0.733 \\ \hline
    CNN + Neg Tag + Dense + Lab/Demo  & 0.893  & 0.822  &  0.749 \\ \hline \hline
    BiLSTM & 0.869 & 0.807 & 0.738 \\ \hline
    BiLSTM + Neg Tag & 0.875 & 0.811 & 0.745 \\ \hline
    BiLSTM + Neg Tag + Dense & 0.892 & 0.823 & 0.739 \\ \hline
    BiLSTM + Neg Tag + Dense + Lab/Demo   & \textbf{0.900} & \textbf{0.833}  & \textbf{0.753}  \\ \hline \hline
    Enc CNN-LSTM &  0.859 & 0.797 & 0.727 \\ \hline
    Enc CNN-LSTM + Lab/Demo &  0.885 & 0.812 & 0.740 \\ \hline
    \end{tabular}
\end{table}

\begin{figure}[h]
\begin{center}
\includegraphics[width=0.7\linewidth]{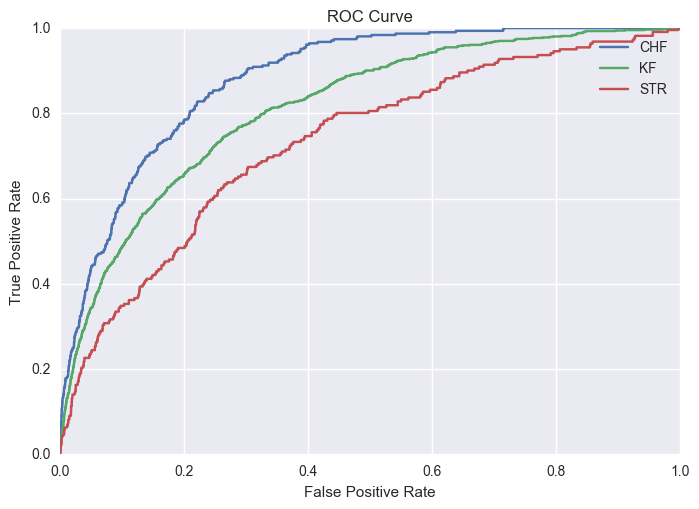}

\end{center}
   \caption{ROC curve of best model on the test set for Congestive Heart Failure (CHF), Kidney Failure (KF), and Stroke (STR) prediction}
\label{fig:roc}
\end{figure}

 The prediction task is highly imbalanced so we report Area under ROC curve (AUC) and precision/recall as the performance measures. We observe that deep learning models with notes outperform all baseline models by large margin. It indicates that notes contain additional information over structured data (comparing to Logistic Reg Lab/Demo and LSTM Lab/Demo) and deep learning model is a promising approach to extract those information (comparing to Logistic Reg Notes). We also find that adding demographics and lab values, as well as negation further improves model performance. The BiLSTM model with negation tagging, an additional dense layer, and the lab and demographic features performs the best across all disease prediction tasks. Figure \ref{fig:roc}  shows the ROC-curve of the best model on each outcome. We achieve relatively high AUCs for all three task. 

In terms of precision and recall, at .15 recall the model achieves a precision of 0.145, 0.152, and 0.025 for CHF, KF, and stroke respectively. At .05 recall the model achieves a precision of 0.255, 0.297, and 0.103. At .01 recall the model achieves a precision of 0.227, 0.183, and 0.053.


\subsection{Explainability and Visualization}

Explainability is crucial for medical predictions as it helps to build trust with medical professionals, improves physician/patient communication and adds some level of transparency of the model behavior. In addition to reporting prediction performance, we present a number of visualization methods for our deep learning models, and compare their effectiveness.

\subsubsection{N-gram importance}

To show the words or phrases that are most relevant to the prediction, we assign a relative importance score to each word based on their impact on the model prediction and highlight the text accordingly. We primarily test the following two approaches to generate the importance score.

\begin{figure}[h!]
\begin{center}
\includegraphics[width=0.85\linewidth]{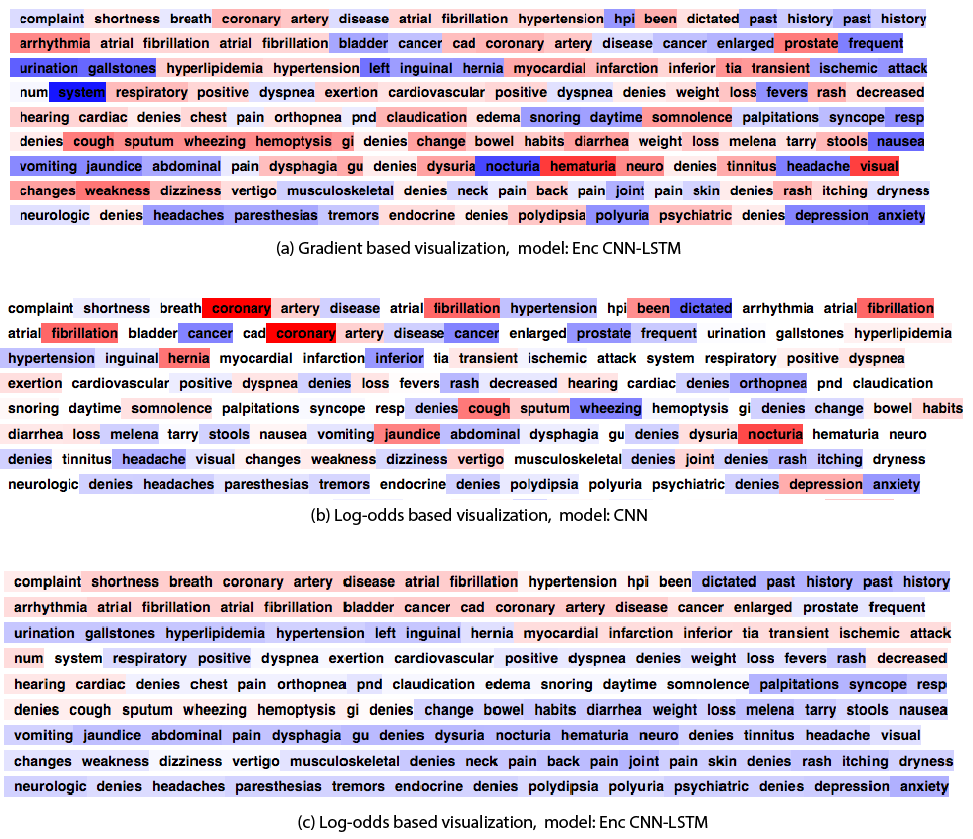}
\end{center}
   \caption{Visualization of N-gram importance comparison. Blue background indicates negative impact on prediction score, i.e., less risky to suffer from heart attack in future; while red background indicates positive impact or increased risk. The darker the color is, the higher the impact.}
\label{fig:viz_comp}
\end{figure}

\textbf{Gradient based approach}

We attempt to measure the level of impact of each word on final prediction by the magnitude of its gradient. We calculate the gradient of the prediction w.r.t. each word embedding and compute the norm. We further divide the norm by the number of occurrence of the word, otherwise tokens like padding has high gradient simply because it appears multiple times. However, we find that this approach to be noisy and not able to generate meaningful interpretations, as shown in Figure \ref{fig:viz_comp} (a). 

\textbf{Log-odds based approach}

The CNN model (Section \ref{sec:CNN}) provides a very transparent and sparse representation of n-grams that contribute to a prediction: those that activate neurons in the max-pooling layer. To show how much each n-gram influences the decision, we use the log-odds that the n-gram contributes to the sigmoid decision function. 

Similar approach can be applied to other type of models, with some modification. For example, in the encounter CNN-LSTM model, we can also identify N-grams that are activated in the CNN max-pooling layer. Then we calculate the log-odds using each individual N-gram as the only input, centered by subtracting the log-odds with only padding tokens as the input. For simplicity, if a word appears in different N-grams of different length, we choose the log-odds based on the longest N-grams. Among N-grams with the same length, we choose the one with highest absolute value. 

Figure \ref{fig:viz_comp} (b) and (c) are examples based on CNN and the encounter model, respectively. Both models highlight some common features such as coronary and fibrillation, while disagree on features such as jaundice and nocturia. We note that features highlighted by CNN are more sparse.

\subsubsection{Convolutional Global Features}

With the log-odds approach, we see that the CNN model is able to identify meaningful features across samples. When examining feature importance across the whole dataset we found that for congestive heart failure, the model is able to learn to use anatomical descriptions of the heart (e.g., artery, diastolic, etc.), words related to heart disease (e.g., hypertension, fibrillation, etc.), heart medication (e.g., furosemide, valsartan, etc.), as well as secondary risk factors (e.g., smoking, alcohol, etc.).

The model also learns to use some less intuitive features. For instance, we found that the word 'daughter' is a relatively strong predictor for heart failure. Upon further investigation we found that the word normally occurs in the context of a patients daughter bringing them in for an appointment which we believe is correlated with age and poor health. 

For more details, Figure \ref{fig:actFeature} in the Appendix shows the average log odds of the top features across the positive examples in the test set by each target.



\section{Related Work and Discussion} 

\subsection{Related work}
Multiple data sources and modeling approaches have been explored to predict future clinical events from historical EHR data. \cite{krishnan2013early} formulated a series of experiments with varying length of patient history window and prediction window to predict diabetes from EHR claims data with regularized logistic regression. \cite{tran2016preterm} predicted preterm births from medication and procedure related features with regularized logistic regression and boosting models. For model structures, deep learning methods have been empirically proven to excel in providing general methods to integrate time series and structured covariates in clinical information. Specifically, CNN and LSTM are apt at encoding temporal information, while Elman RNN and LSTM were shown to perform well for vital signs [\cite{graves2013speech}]. Deep CNNs perform well for clinical diagnostics [\cite{razavian2015temporal}]. \cite{baumel2017multi} showed hierarchical model with attention and GRU cells works well with discharge notes, although their task is ICD assignment of current encounter instead of prediction of future events. 

For integrating mix-type inputs, \cite{fiterau2017shortfuse} introduced methods to directly model interaction between static information (e.g. age, gender, etc.) and sequential inputs. In CNN, the static covariates are provided as parameters to the convolution function across time; and in LSTM, the static covariates are added to the input of the gates functions, although directly concatenating static covariates to sequential covariates were argued inefficient for learning and prone to overfitting. \cite{sureshclinical} predicted clinical intervention combining structured data and notes. Clinical narrative notes were transformed to a 50-dimensional vector of topic proportions for each note using Latent Dirichlet Allocation, and static variables were replicated across time. All inputs were then concatenated. Their empirical results showed LSTM / CNN based model outperformed logistic regression baseline with the same input by large margins. \cite{tran2016preterm} also used free-text data in their work, although the authors simply extracted uni-grams from the notes after removing stop-words. 

For auxiliary information, \cite{choi2017gram} utilized a hierarchical structure from the medical term ontology via an attention mechanism. The motivation is to deal with small sample issue for certain low-level codes. Though their empirical result seems to show that in certain tasks, simply rolling up to the higher level codes or RNN with GloVe embeddings can have comparable performances.

For clinical text representation learning,  \cite{moen2013distributional} trained a set of word2vec embedding on PubMed data. Moreover, \cite{wu2017starspace} introduced StarSpace as a general purpose representation learning framework. Entities are represented by the sum of their features' embeddings, which are trained by optimizing a loss function that compares similar entity pairs with sampled negative pairs. Users will define the 'label' to determine similarity, e.g. sentence and the article topic is a similar pair while with other topics are negative pairs. The Information Retrieval task turns out to be most relevant to the note representation purpose. 

\subsection{Discussion and Future Directions}
In this work we have evaluated a wide variety of architectures for preventable disease prediction as well as a range of variations for adapting these models better to the problem. With a combination of medical notes, structured values and our novel negation tagging technique we are able to achieve very strong predictive accuracy in terms of AUC. We have shown that our model outperforms the use of only structured numerical data as well as standard TF-IDF techniques. While these results are promising, our model needs to achieve a higher precision to be actionable when deployed in the clinical setting. We may further improve the model in future iterations by incorporating other sources of structured data and more sophisticated model architectures. Additionally, we identify the following potential exploration directions:

\textbf{Quantitative Variables}

In this work we use a simple yet efficient regular expression tool [\cite{hao2016valx}] to extract the numerical features and it was able to increase model performance. However, we note that accuracy of the regular expression tool is a bottleneck for these features, especially in terms of associating the value to the right test item. A more sophisticated extraction method could potentially improve predictive performance. 

\textbf{Deep Learning Negation Tagging}

Motivated by our successful use of negation tags, we want to introduce a more sophisticated negation tagger. Negex is very efficient however there are different deep learning architectures that can tag negations with higher precision and recall, and be trained along with the rest of the task, end-to-end. We expect this to improve model performance. 

\textbf{Additional Prediction Tasks}

In this work we selected the 3 disease areas (congestive heart failure, kidney failure, stroke) and the feature-gap-prediction time window based on data availability and suggestions from clinical experts. It would be interesting to evaluate and compare model performance in additional disease areas, with various levels of data availability and time-windows of different lengths. 

\textbf{Alternative Visualization Approaches}

We are also interested in exploring alternative visualization approaches in addition to improving model performance. For example, we can mask individual features and evaluate change in prediction to take better consideration of interactions between features. Alternatively, we can optimize a masking function over inputs to evaluate relative importance. 


\acks{We would like to express special thanks to Yin Aphinyanaphongs, Marina Marin, Himanshu Grover at NYUMC Predictive Analytics Unit, Michael Cantor and the NYUMC Data Core Team for EHR data access, retrieval, anonymization and helpful feedback; and Jennifer Scherer for advices on disease definitions. We also want to thank the NYUMC HPC team and Joe Katsnelson for providing GPU access and technical supports, and Sung Pil Moon for visualization tool prototyping. We are also grateful to Claudio Silva and Viola Cao for feedbacks on early versions of the work, and to Leora Horwit, Yin Aphinyanaphongs and anonymous reviewers of MLHC for their insightful comments on the paper draft.}



\newpage
\bibliography{sample}

\newpage
\appendix
\section{Appendix}

\subsection{Target disease mapping to ICD-10 codes}
\label{sec:icd10}
\textbf{Congestive heart failure: }
I09.81, I11.0, I13.0, I13.2, I50.1, I50.20, I50.21, I50.22, I50.23, I50.30, I50.31, I50.32, I50.33, I50.40, I50.41, I50.42, I50.43, I50.9, I09.81, I50.1, I50.20, I50.21, I50.22, I50.23, I50.30, I50.31, I50.32, I50.33, I50.40, I50.41, I50.42, I50.43, I50.810, I50.811, I50.812, I50.813, I50.814, I50.82, I50.83, I50.84, I50.89, I50.9\\

\textbf{Kidney failure: }
A18.11, A52.75, B52.0, C64.1, C64.2, C64.9, C68.9, D30.00, D30.01, D30.02, D41.00, D41.01, D41.02, D41.10, D41.11, D41.12, D41.20, D41.21, D41.22, D59.3, E08.21, E08.22, E08.29, E08.65, E09.21, E09.22, E09.29, E10.21, E10.22, E10.29, E10.65, E11.21, E11.22, E11.29, E11.65, E13.21, E13.22, E13.29, E74.8, I12.0, I13.11, I13.2, I70.1, I72.2, K76.7, M10.30, M10.311, M10.312, M10.319, M10.321, M10.322, M10.329, M10.331, M10.332, M10.339, M10.341, M10.342, M10.349, M10.351, M10.352, M10.359, M10.361, M10.362, M10.369, M10.371, M10.372, M10.379, M10.38, M10.39, M32.14, M32.15, M35.04, N00.0, N00.1, N00.2, N00.3, N00.4, N00.5, N00.6, N00.7, N00.8, N00.9, N01.0, N01.1, N01.2, N01.3, N01.4, N01.5, N01.6, N01.7, N01.8, N01.9, N02.0, N02.1, N02.2, N02.3, N02.4, N02.5, N02.6, N02.7, N02.8, N02.9, N03.0, N03.1, N03.2, N03.3, N03.4, N03.5, N03.6, N03.7, N03.8, N03.9, N04.0, N04.1, N04.2, N04.3, N04.4, N04.5, N04.6, N04.7, N04.8, N04.9, N05.0, N05.1, N05.2, N05.3, N05.4, N05.5, N05.6, N05.7, N05.8, N05.9, N06.0, N06.1, N06.2, N06.3, N06.4, N06.5, N06.6, N06.7, N06.8, N06.9, N07.0, N07.1, N07.2, N07.3, N07.4, N07.5, N07.6, N07.7, N07.8, N07.9, N08, N13.1, N13.2, N13.30, N13.39, N14.0, N14.1, N14.2, N14.3, N14.4, N15.0, N15.8, N15.9, N16, N17.0, N17.1, N17.2, N17.8, N17.9, N18.1, N18.2, N18.3, N18.4, N18.5, N18.6, N18.9, N19, N25.0, N25.1, N25.81, N25.89, N25.9, N26.1, N26.9, Q61.02, Q61.11, Q61.19, Q61.2, Q61.3, Q61.4, Q61.5, Q61.8, Q62.0, Q62.2, Q62.10, Q62.11, Q62.12, Q62.31, Q62.32, Q62.39, R94.4\\

\textbf{Stroke: }
G45.0, G45.1, G45.2, G45.8, G45.9, G46.0, G46.1, G46.2, G97.31, G97.32, I60.00, I60.01, I60.02, I60.10, I60.11, I60.12, I60.20, I60.21, I60.22, I60.30, I60.31, I60.32, I60.4, I60.50, I60.51, I60.52, I60.6, I60.7, I60.8, I60.9, I61.0, I61.1, I61.2, I61.3, I61.4, I61.5, I61.6, I61.8, I61.9, I63.00, I63.02, I63.011, I63.012, I63.019, I63.031, I63.032, I63.039, I63.09, I63.10, I63.111, I63.112, I63.119, I63.12, I63.131, I63.132, I63.139, I63.19, I63.20, I63.211, I63.212, I63.219, I63.22, I63.231, I63.232, I63.239, I63.29, I63.30, I63.311, I63.312, I63.319, I63.321, I63.322, I63.329, I63.331, I63.332, I63.339, I63.341, I63.342, I63.349, I63.39, I63.40, I63.411, I63.412, I63.419, I63.421, I63.422, I63.429, I63.431, I63.432, I63.439, I63.441, I63.442, I63.449, I63.49, I63.50, I63.511, I63.512, I63.519, I63.521, I63.522, I63.529, I63.531, I63.532, I63.539, I63.541, I63.542, I63.549, I63.59, I63.6, I63.8, I63.9, I66.01, I66.02, I66.03, I66.09, I66.11, I66.12, I66.13, I66.19, I66.21, I66.22, I66.23, I66.29, I66.3, I66.8, I66.9, I67.841, I67.848, I67.89, I97.810, I97.811, I97.820, I97.821

\newpage
\subsection{Note summary statistics}

\begin{figure}[H]
    \centering
    \includegraphics[width=.8\textwidth]{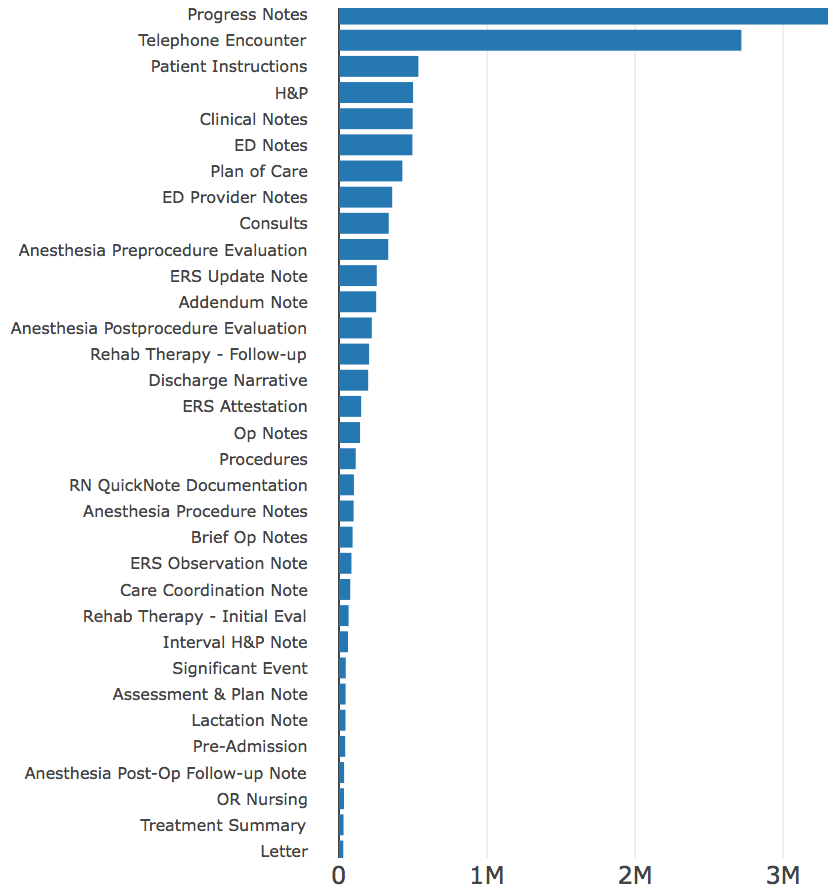}
    \caption{Note Type Distribution}\label{fig:notetype1}
\end{figure}

\begin{figure}[H]
    \centering
    \includegraphics[width=0.8\textwidth]{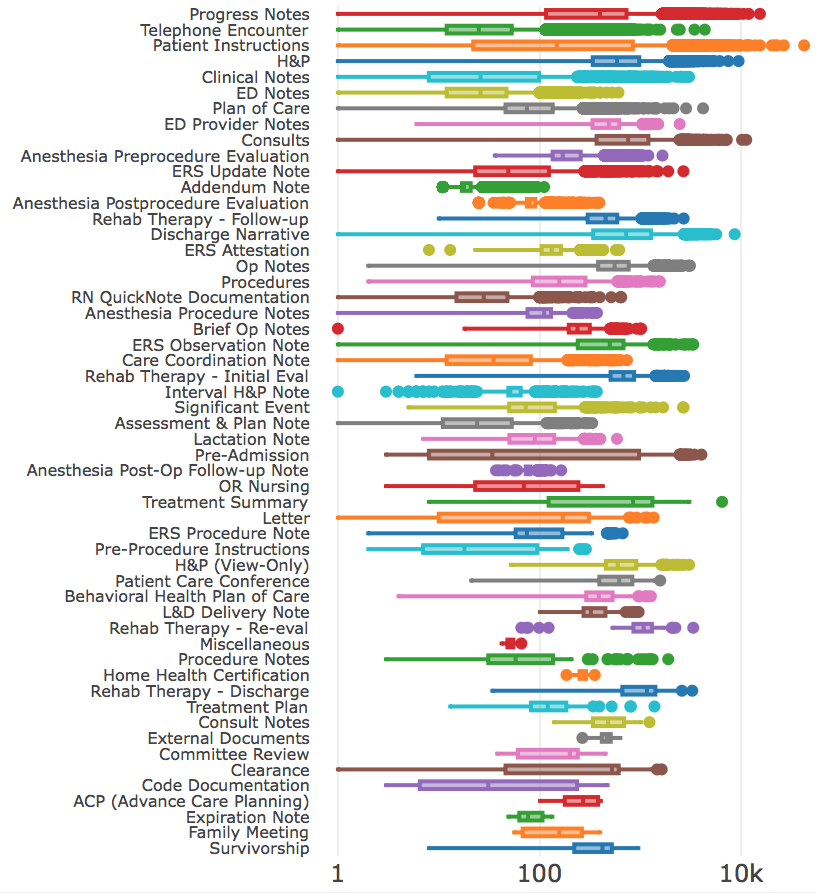}
    \caption{Number of Words (log-scale) Per Note Distribution}\label{fig:notetype2}
\end{figure}

\newpage

\subsection{Patient demographics}

\begin{figure}[h]
    \centering
    \includegraphics[width=.7\textwidth]{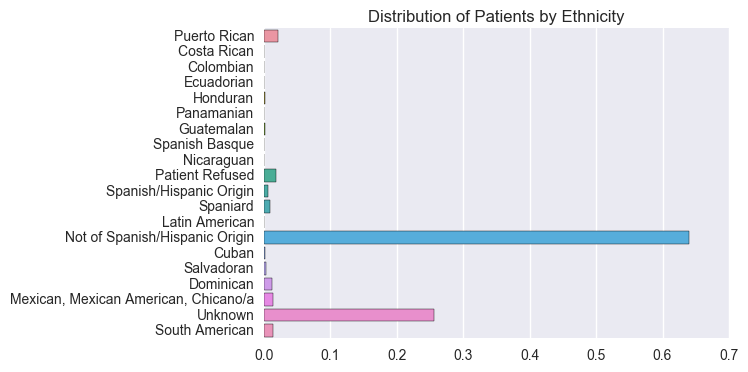}
    \caption{Distribution of patients by ethnicity}\label{fig:distethnicity}
\end{figure}

\begin{figure}[h]
    \centering
    \includegraphics[width=.7\textwidth]{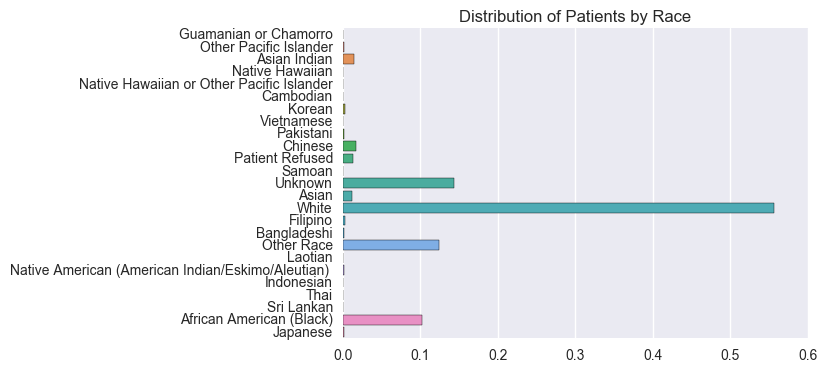}
    \caption{Distribution of patients by race}\label{fig:distrace}
\end{figure}

\begin{figure}[h]
    \centering
    \includegraphics[width=.5\textwidth]{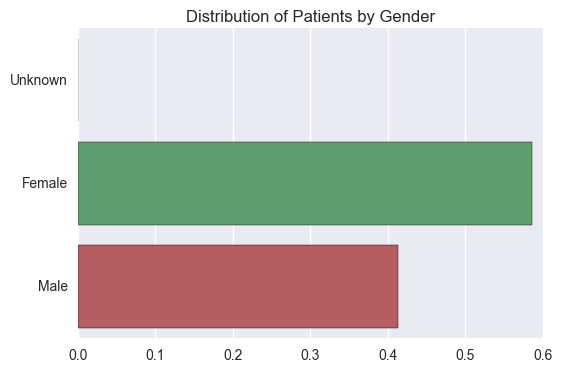}
    \caption{Distribution of patients by gender}\label{fig:distgender}
\end{figure}

\newpage

\subsection{Lab values prevalence}
\begin{table}[h]
    \centering
    \caption{Top 30 most frequent lab values}
\begin{tabular}{ l  c }
\hline
Item name & Prevalence by percentage of encounters \\ \hline
Weight & $65.3\%$ \\
BP\_systolic & $61.5\%$ \\
BP\_diastolic & $61.4\%$ \\
Height & $55.2\%$ \\
Pulse & $54.7\%$ \\
Oxygen saturation & $38.2\%$ \\
Temperature & $37.9\%$ \\
Resp & $29.3\%$ \\
BMI & $27.7\%$ \\
Urea Nitrogen & $17.0\%$ \\
Creatinine & $16.7\%$ \\
Chloride & $15.3\%$ \\
Potassium & $15.1\%$ \\
Sodium & $15.0\%$ \\
Carbon Dioxide & $14.6\%$ \\
Hemoglobin & $14.0\%$ \\
Hematocrit & $13.3\%$ \\
Glucose & $13.3\%$ \\
Alanine Aminotransferase & $12.5\%$ \\
Aspartate Aminotransferase & $12.3\%$ \\
Ery. Mean Corpuscular Volume & $12.2\%$ \\
Alkaline Phosphatase & $10.7\%$ \\
Bilirubin & $10.6\%$ \\
Platelets & $10.5\%$ \\
Calcium & $10.3\%$ \\
Leukocytes & $5.7\%$ \\
WBC & $5.5\%$ \\
HDL Cholesterol & $5.3\%$ \\
LDL Cholesterol & $3.9\%$ \\
Albumin & $3.9\%$ \\ \hline
\end{tabular}

\label{table:labs}
\end{table}

\newpage

\subsection{Network Architecture Details}
\label{sec:architecture}

\textbf{\quad \ \ LSTM baseline with lab/demographics}

1 layer LSTM with GRU cells and 256 hidden nodes. 

Two dense layers with ReLU transformation, 512 and 256 hidden nodes respectively before output layer. Batch normalization applied.

Dropout with ratio of .5 applied to all levels of network. \\

\textbf{CNN}

Convolutional layer has kernel size 1,2, and 3, 256 hidden units and ReLU activation. 

Dropout with ratio of .3 and batch normalization are applied at all levels of the network.\\

\textbf{CNN + dense}

Same structure as CNN with an additional dense hidden layer with 256 units before output layer.\\

\textbf{BiLSTM}

1 layer BiLSTM with 256 hidden units in each direction. 

We do not use dropout or batch normalization in this model.\\

\textbf{BiLSTM + dense}

Same structure as BiLSTM with an additional dense hidden layer with 256 hidden units before output layer.  \\

\textbf{Encounter CNN-LSTM}

Convolution layer has kernel size 1 to 5, 128 hidden units, ReLU activation and batch normalization. 

1 layer LSTM with Gated Recurrent Unit (GRU) and 256 hidden units. 

An dense hidden layer with 256 units before output layer. Batch normalization applied. 

Dropout with ratio of .5 is applied at all levels of the network.\\

\textbf{Encounter CNN-LSTM with lab and demographics}

Convolution layer has kernel size 1 to 5, 128 hidden units, ReLU activation and batch normalization. 

1 layer LSTM with Gated Recurrent Unit (GRU) and 512 hidden units. 

An dense hidden layer with 256 units before output layer. Batch normalization applied. 

Dropout with ratio of .5 is applied at all levels of the network.

\subsection{Feature contribution}
\begin{figure}[h]
    \centering
    \begin{subfigure}[b]{0.85\textwidth}
        \includegraphics[width=\textwidth]{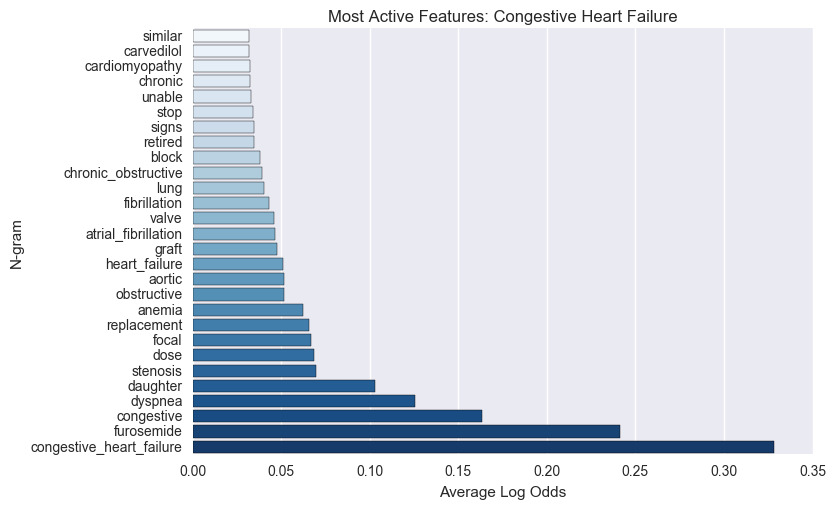}
    \end{subfigure}
    \begin{subfigure}[b]{0.85\textwidth}
        \includegraphics[width=\textwidth]{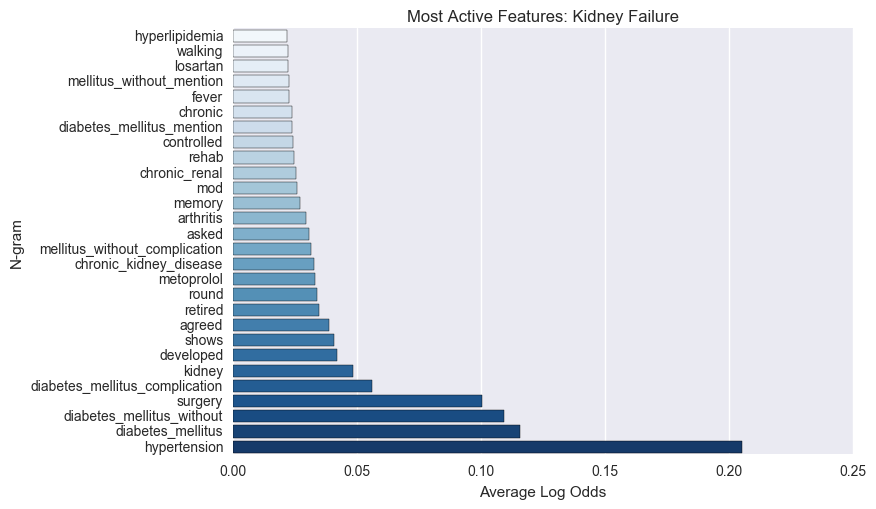}
    \end{subfigure}
    \begin{subfigure}[b]{0.85\textwidth}
        \includegraphics[width=\textwidth]{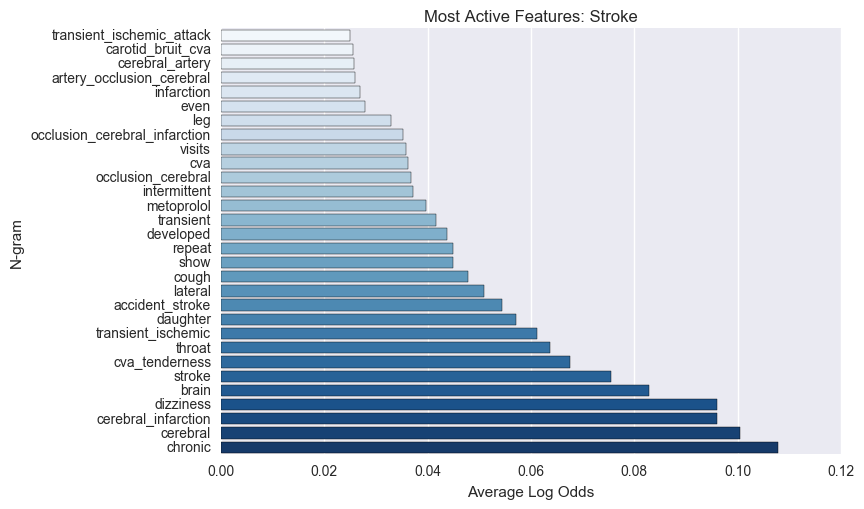}
    \end{subfigure}
    
    \caption{Average log odds that each feature contributes to the sigmoid decision function, CNN model}
    \label{fig:actFeature}
\end{figure}


\end{document}